\pdfoutput=1

\documentclass[11pt]{article}

\usepackage{array}
\usepackage[table]{xcolor}
\usepackage{fix-cm}

\usepackage[preprint]{acl}
\usepackage{subcaption} 
\usepackage{graphicx}
\usepackage{times}
\usepackage{latexsym}
\usepackage{multicol, blindtext}
\usepackage{amsmath} 
\usepackage{booktabs}
\usepackage{multirow}
\usepackage[T1]{fontenc}
\usepackage{longtable}
\usepackage{todonotes}
\usepackage[utf8]{inputenc}

\usepackage{microtype}

\usepackage{inconsolata}
\usepackage{tikz}
\usetikzlibrary{arrows,calc,positioning}
\usepackage{graphicx}
\usepackage{float}
\usepackage{amssymb}
\usepackage{stfloats}
\usepackage{enumitem}
\usepackage{tabularx}

\newcommand{\vehiclecolor}{\textcolor{red}}
\newcommand{\loadcolor}{\textcolor{blue}}
\newcommand{\finecolor}{\textcolor{green!50!black}}
\newcommand{\namecolor}{\textcolor{magenta}}
\newcommand{\damagecolor}{\textcolor{cyan}}
\newcommand{\locationcolor}{\textcolor{brown}}
\newcommand{\outcomecolor}{\textcolor{orange!80!black}}
\newcommand{\newdetailcolor}{\textcolor{violet}}
\newtheorem{hyp}{Hypothesis}
\title{LLM as a Broken Telephone: Iterative Generation Distorts Information}

\author{
 \textbf{Amr Mohamed\textsuperscript{1}$^\dagger$},
 \textbf{Mingmeng Geng\textsuperscript{2}},
 \textbf{Michalis Vazirgiannis\textsuperscript{1,3}},
 \textbf{Guokan Shang\textsuperscript{1}$^\dagger$}
\\
\\
 \textsuperscript{1}MBZUAI,
 \textsuperscript{2}SISSA,
 \textsuperscript{3}Ecole Polytechnique
\\
 \small{
$^\dagger$Correspondence: \texttt{\{amr.mohamed, guokan.shang\}@mbzuai.ac.ae}
 }
}

\begin{document}
\maketitle
\begin{abstract}
As large language models are increasingly responsible for online content, concerns arise about the impact of repeatedly processing their own outputs.
Inspired by the ``broken telephone'' effect in chained human communication, this study investigates whether LLMs similarly distort information through iterative generation.
Through translation-based experiments, we find that distortion accumulates over time, influenced by language choice and chain complexity. 
While degradation is inevitable, it can be mitigated through strategic prompting techniques. 
These findings contribute to discussions on the long-term effects of AI-mediated information propagation, raising important questions about the reliability of LLM-generated content in iterative workflows.
\end{abstract}

\section{Introduction}
Large Language Models (LLMs) are becoming an integral part of our daily lives, helping us process, comprehend, and convey information via text, while also expanding their support to additional areas \citep{yin2023survey}.
Consequently, an increasing amount of online content is now model-generated or assisted \citep{geng2024chatgpt}, and such content is almost indistinguishable from human-produced data \citep{uchendu2023does}.

This prompts us to consider the question: what effects arise when the same piece of information is repeatedly processed by LLMs through multiple iterations?
This procedure is analogous to the telephone game in human communication, a widely known children’s game in which a message is passed sequentially from one player to the next, with the final version often differing significantly from the original, usually with amusing or humorous effect.
This happens because players often act as \textit{broken telephones}, where information is gradually distorted as it is passed along the chain of individuals, highlighting how repeated transmission can lead to the accumulation of errors, omissions, or unintended alterations \citep{hitchcock2011information}.

Investigating these effects for LLMs is becoming increasingly crucial in the present era, because LLMs are not only consuming human-supplied information at one time, but also processing their own outputs in an iterative way.
Therefore, our study focuses on exploring whether LLM also acts as a broken telephone, when the same content is continuously refined, paraphrased, or reprocessed, and particularly when the generated output becomes the input for subsequent model iterations. 
We expect to observe an effect similar to that of human information distortion through iterative generation.

\begin{table}[t]
    \centering
    \renewcommand{\arraystretch}{1.2} 
    \setlength{\tabcolsep}{2pt} 
    \begin{tabularx}{\columnwidth}{c X}
        \small 0$^{th}$ & \small A \vehiclecolor{lorry} driver has been \finecolor{fined} after his load of \loadcolor{slabs} fell off his vehicle on a bend, writing off a passing car worth £50,000.\\
        \hline
        \small 2$^{nd}$  & \small A \vehiclecolor{lorry} driver was \finecolor{fined} after a \loadcolor{stone slab} he was transporting fell off his vehicle at a bend, causing damage to a passing car worth up to \underline{£50,000}. \\
        \hline
        \small 10$^{th}$  & \small A \vehiclecolor{bus} received a \underline{\$50,000} \finecolor{fine} after a \loadcolor{large rock} dislodged from it at a bend and forced passing cars to swerve off the road. \\
        \hline
        \small 50$^{th}$  & \small A \vehiclecolor{bus} received a \finecolor{compensation} of \textbf{\$}50,000 after a \loadcolor{large rock} struck the bus when the bus changed lanes on the city road, \newdetailcolor{causing damage and an explosion on the road}.\\
        \hline
        \small 100$^{th}$  & \small A \vehiclecolor{small car} received a \finecolor{compensation} of \$50,000 after a \loadcolor{large rock} collided with the car, \newdetailcolor{causing an accident and an explosion on the road}.\\
    \end{tabularx}
    \caption{Example of iterative translations of an English news article using \textit{Llama-3.1-8B-Instruct}, with Thai as the intermediate language, highlighting the distortions introduced over the different iterations.} 
    \label{tab_eg}
\end{table}


In our study, we simulate the LLMs' telephone game primarily through the task of translation, as iterative translation serves as a critical and tangible testbed for examining how meaning and form degrade over repeated generations. This setup reflects real-world scenarios—for instance, cross-lingual news transmission—where content is repeatedly translated across languages. The extent of distortion, shaped by the interplay between a language's representation in the training data and its linguistic similarity to the source, is not unique to translation but illustrates a broader phenomenon also observed in other iterative LLM tasks, such as rephrasing, which we investigate under three experimental setups.

\begin{figure*}[t]
    \centering
    \includegraphics[width=\textwidth]{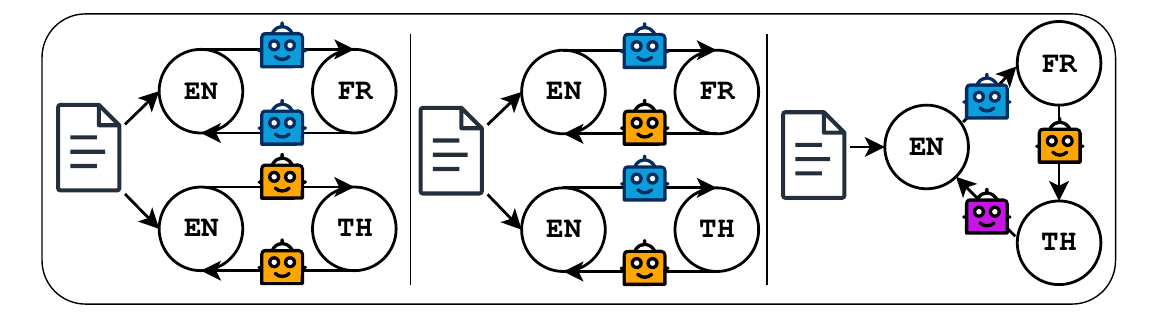}
    \caption{Overview of examples from our three experimental setups. \textbf{\textit{Left}}: Bilingual Self-Loop---A single model iteratively translates a document from English (EN) to French (FR) or Thai (TH) and back to English. \textbf{\textit{Middle}}: Bilingual Two-Player---Two different models collaborate within the same chain on translating between English and French or English and Thai. \textbf{\textit{Right}}: Multilingual Multiplayer---A more complex translation chain involving multiple languages and models, designed to examine how increasing the variety of languages and models accelerates information distortion over iterative generations.}
    \label{fig:experiments_illustration}
\end{figure*}

As illustrated in Figure \ref{fig:experiments_illustration},
within each iteration, a document in English is subsequently translated into one or more different languages, then back to English, by leveraging LLMs.
We compare the back-translated English version with the initial English version at every iteration with textual relevance and factuality measures, to investigate whether and how information distortion accumulates.
Our results show that over time, small alterations in phrasing, meaning, or factual details can accumulate, leading to a progressive drift from the original source, as illustrated by the example in Table \ref{tab_eg}. Code and data are publicly available\footnote{\url{https://github.com/amr-mohamedd/LLM-as-a-Broken-Telephone}}. Our main findings include:

\noindent$\bullet$ The degree of information distortion in translation chains depends on the choice of intermediate languages, influenced by their linguistic similarity to the source language and their prevalence in the model's pre-training and post-training corpora.  

\noindent$\bullet$ Greater chain complexity, whether by adding languages or models, often amplifies distortion, with longer chains introducing more degradation regardless of the type of iterative chain.  

\noindent$\bullet$ Although distortion is unavoidable, it can be mitigated through temperature control and restricted prompting, which restrict the LLM from deviating significantly from the original text.
    
Our research echoes the ongoing conversation about the long-term impact of the widespread use of LLM-generated content on models themselves, humans, and society at large---often termed \textit{model} and \textit{knowledge} collapse \citep{guo-etal-2024-curious,peterson2024ai}.
Our findings raise concerns about the reliability of AI-mediated information dissemination over the long term and in an iterative way.

\section{Related Work}
\noindent\textbf{Model Collapse.} Iterative training on synthetically generated data induces model collapse, a phenomenon characterized by systematic erosion of the long-tail components of the original data distribution \citep{shumailov2023curse}. Theoretical analyses further elucidated how self-consuming training loops alter intrinsic scaling laws, thereby intensifying this collapse \citep{fu2024towards,dohmatob2024tale}, complementing earlier findings on distributional distortions \citep{lebrun2022evaluating}. Furthermore, \citet{guo-etal-2024-curious} demonstrated that iterative training on synthetic text does not preserve the nuanced richness of human language, particularly in creative tasks, underscoring the broader challenges of maintaining linguistic diversity in iteratively generated content.\\\vspace{-8pt}

\noindent\textbf{Iterative Generation and Information Evolution.} Iterative generation can trigger model collapse, whereby the diversity of real-world information degrades over time—a process that \citet{peterson2024ai} defines as knowledge collapse.  Research on language evolution offers a framework for analyzing these degradations \citep{markov2023language}, aligning with broader perspectives on cultural evolution \citep{MesoudiWhiten2008,CaldwellMillen2008}. 
In the context of LLMs, \citet{perez2024llms} analyzed text properties evolution in rephrasing, continuation, and inspiration-taking tasks. Their work, however, overlooked translation—a key LLM application—and focused solely on chains involving a single model. 
Our work overcomes these shortcomings by investigating how iterative information translation accelerates distortions, explores heterogeneous model chains, and extends the analysis to higher complexity rephrasing chains, providing a broader view of iterative generation's impact on information evolution.\\\vspace{-9pt}

\noindent\textbf{LLM Agents.} We consider the implications for multi-agent settings, where communication frameworks leverage collaborative interactions between multiple LLMs \citep{park2023generative, Wu2023-nb,Li2024-yw}. These frameworks enable agents to iteratively refine outputs through debate-style interactions \citep{Helm2024-yh} or cooperative task decomposition \citep{Pham2023-al}, often improving accuracy in mathematical and logical tasks \citep{zhang-etal-2024-exploring}. As introduced by \citet{park2023generative}, generative agents showcase the potential for creating interactive simulacra of human behavior through memory, reflection, and planning. However, such architectures implicitly assume that iterative exchanges preserve or enhance information fidelity—a premise challenged by our findings in translation chains. While prior work focuses on emergent problem-solving capabilities \citep{chan2024scaling}, our study reveals how these same iterative mechanisms accelerate information distortion, particularly in scenarios where translation ambiguities compound through successive agent handoffs.\\\vspace{-9pt}

\noindent\textbf{Evaluation of LLM Outputs.} In addition to the multi-agent perspective, it is essential to scrutinize how LLM outputs are evaluated. Existing research predominantly relies on metrics such as token similarity \citep{Hu2024UnveilingLE}, output diversity \citep{guo2024benchmarking,shaib2024standardizing}, and factuality \citep{Wang2023-lo, Iqbal2024-vd, Min2023-ls, Chern2023-xl}. However, these evaluations are generally confined to single iterations and fail to capture the cumulative degradation introduced by iterative generation—a critical aspect of the translation chains under investigation. Although previous studies have explored variations in toxicity, positivity, difficulty, and length in iterative LLM transmission chains \citep{perez2024llms}, they have overlooked the systematic assessment of textual similarity and factuality. Our work addresses this gap by providing a rigorous analysis of the deterioration of these properties over successive iterations in both translation and rephrasing tasks.\vspace{-9pt}

\section{Methodology}
\vspace{-5pt}
In this section, we formalize the telephone game procedure with machine translation, noting that the broken telephone effect may occur with any generative task when carried out iteratively.

\subsection{Notations and Definitions}
\label{sec:notation}
Let $\mathcal{D} = \{d_i\}_{i=1}^{I}$ denote a set of $I$ \textit{documents}, $\mathcal{L} = \{l_j\}_{j=1}^{J}$ as a set of $J$ natural \textit{languages}, and $\mathcal{M} = \{m_k\}_{k=1}^{K}$ for a set of $K$ \textit{models}. 

We define a \emph{translation chain} as a sequence of \(N\) translation iterations that progressively transform a document. For iteration \(t \ge 1\), let \(d_{i,l_{\text{source}}}^{(t-1)}\) be the \(i\)-th document in the source language at iteration \(t-1\).
At iteration \(t\), an ordered language chain is constructed by selecting a permutation \(\pi^{(t)}\) of \(J-1\) languages from \(\mathcal{L}\) and forming the sequence
\begin{equation}
    \mathcal{L}^{(t)} = (l_1^{(t)},\, l_2^{(t)},\, \dots,\, l_{J-1}^{(t)},\, l_J^{(t)})
\end{equation}
with the requirement that \(l_J^{(t)} = l_{\text{source}}\) (ensuring that the final translation returns to the source language). Simultaneously, a model sequence 
\begin{equation}
    \mathcal{M}^{(t)} = \bigl(m_1^{(t)},\, m_2^{(t)},\, \dots,\, m_{J}^{(t)}\bigr)
\end{equation}
is defined, where each \(m_k^{(t)}\) is sampled uniformly from \(\mathcal{M}\) (allowing repeats; if \(|\mathcal{M}|=K=1\), the same model is used throughout).

Let \(\mathcal{T}_{a \leftarrow b}^{m}(\cdot)\) denote the translation operator that converts an input from language \(b\) to language \(a\) using model \(m\). The composed operator for iteration \(t\) is then 
\begin{equation}
    \mathcal{T}^{(t)} = \mathcal{T}_{l_J^{(t)} \leftarrow l_{J-1}^{(t)}}^{m_J^{(t)}} \circ \cdots \circ \mathcal{T}_{l_2^{(t)} \leftarrow l_1^{(t)}}^{m_2^{(t)}} \circ \mathcal{T}_{l_1^{(t)} \leftarrow l_{\text{source}}}^{m_1^{(t)}}
\end{equation}
so that the updated document is given by
\begin{equation}
  d_{i,l_{\text{source}}}^{(t)} = \mathcal{T}^{(t)}\Bigl(d_{i,l_{\text{source}}}^{(t-1)}\Bigr).  
\end{equation}
Starting with \(d_{i,l_{\text{source}}}^{(0)} = d_i\), the process yields the sequence
$(d_{i,l_{\text{source}}}^{(0)},\, d_{i,l_{\text{source}}}^{(1)},\, \dots,\, d_{i,l_{\text{source}}}^{(N)})$,
where \(N\) is the total number of iterations.

\subsection{Experimental Settings}

\noindent\textbf{Languages.} We selected \textit{English} (\textit{EN}) as $l_{\text{source}}$ for all experiments and \textit{French} (\textit{FR}), \textit{German} (\textit{DE}), \textit{Dutch} (\textit{NL}), \textit{Vietnamese} (\textit{VN}), \textit{Chinese} (\textit{ZH}), and \textit{Thai} (\textit{TH}) as the \textit{bridge} (intermediate) languages in the translation chains.
Within each iteration, a document in English is subsequently translated into one or more bridge languages, then back to English.
This set creates varying degrees of semantic, lexical, and syntactic similarities between the source language and the bridge languages, which may differentially influence the extent of distortion introduced within the translation chains \citep{marchisio-etal-2020-unsupervised,guerin2024impact}.

\noindent\textbf{Datasets.} We utilized three datasets that span distinct domains: \textit{BookSum} \citep{kryscinski2021booksum}, \textit{ScriptBase-alpha} \citep{gorinski-lapata-2015-movie}, and \textit{(BBC)News2024} \citep{li2024latesteval}, from which we select articles published in 2024 to minimize the chances of data exposure that may result in biases amplification over the iterations \citep{luo-etal-2024-diverge, li2024latesteval}. For our experiments, we randomly select 150 documents from each dataset, with each document containing between 100 and 200 words long.

\noindent\textbf{Models.} We primarily used two models, \textsc{Llama-3.1-8B-Instruct} (\textit{Llama}) \citep{dubey2024llama} and \textsc{Mistral-7B-Instruct-v0.2} (\textit{Mistral}) \cite{jiang2023mistral}, for our main experiments. Additionaly, \textsc{Gemma-2-9B-it} (\textit{Gemma}) \citep{team2024gemma} is incorporated into Experiment 3 (Section \ref{sec:exp3}) to evaluate higher complexity chains. 

\noindent\textbf{Decoding Parameters and Translation Prompt.} Each model was used for inference with its default decoding parameters. We capped the maximum number of newly generated tokens at 8000 to encourage open-ended generation. This high limit allows translations, which can vary in length across different languages, to conclude naturally rather than being prematurely truncated. Models within the main experiments were prompted to translate documents from a source to a target language with a moderately constrained prompt. The full translation prompt can be found in Appendix \ref{sec:prompts}.

\subsection{Evaluation Metrics}
To comprehensively assess the impact of iterative generation on text quality, we employ two complementary sets of evaluation metrics: textual relevance and factuality preservation. The former quantifies the lexical, syntactic, and semantic deviations introduced at each generation step, while the latter evaluates the degree to which the generated text remains faithful to the original information. 

\noindent\textbf{Textual Relevance.} We used \textbf{BLEU} \citep{papineni-etal-2002-bleu} to detect incremental errors, \textbf{ROUGE-1} \citep{lin-2004-rouge} to quantify word-level omissions and subtle deviations, \textbf{CHR-F} \citep{popovic-2015-chrf} for capturing character-level deviations and errors accumulation, \textbf{METEOR} \cite{banerjee-lavie-2005-meteor} for being adept at capturing paraphrastic variations and subtle semantic shifts, and finally \textbf{BERTScore} \citep{Zhang2019BERTScoreET} for its focus on nuanced contextual and semantic relationships beyond traditional n-gram overlap-based methods.

\noindent\textbf{Factuality Preservation.} \textbf{FActScore} \citep{Min2023-ls} decomposes long-form text into atomic units and verifies each against a trusted reference using a dedicated judge model.
In this study, we assume that the original text is factually correct and use FActScore to assess the rate of factuality degradation over the different iterations by comparing each model generation with its original text, then employ \textit{Claude 3.5 Sonnet} to be the judge model.\vspace{-2pt}


\section{Experiments}
\label{sec:experiments}
\subsection{Experiment 1: Bilingual Self-loop}
\label{sec:exp1}  

\begin{figure*}[t]
\center
  \includegraphics[width=\textwidth]{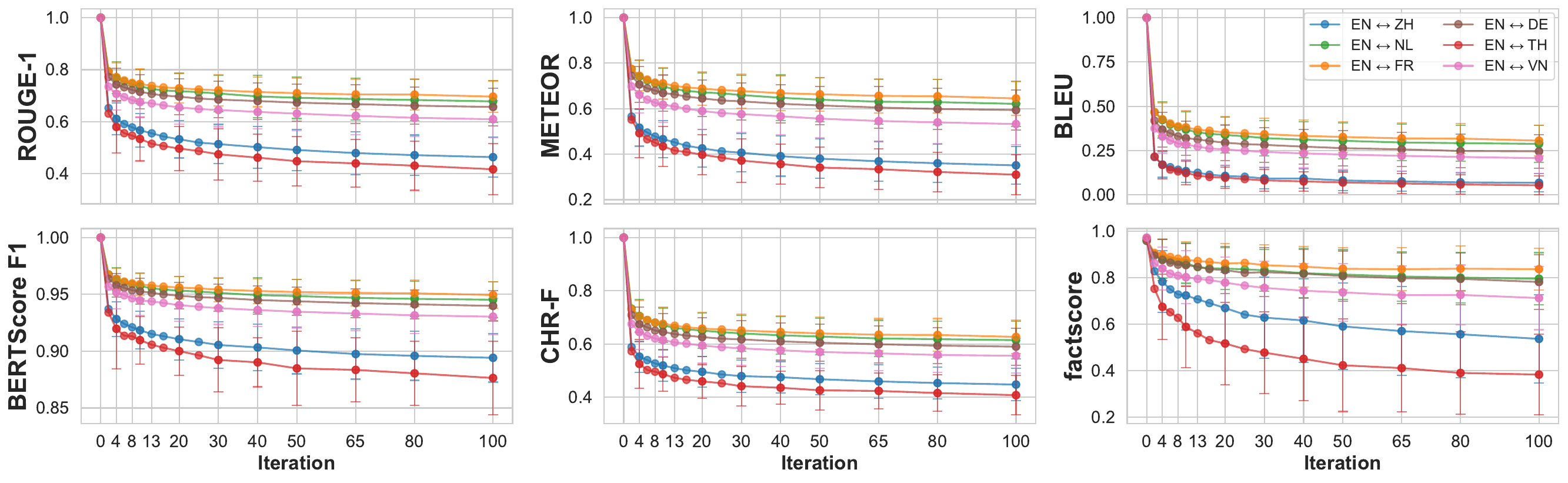}
  \caption{Results of \textit{Llama} in the Bilingual Self-loop Experiment showing metrics evolution across translation iterations over the \textit{News2024} dataset for \textit{French} (\textit{FR}), \textit{German} (\textit{DE}), \textit{Dutch} (\textit{NL}), \textit{Vietnamese} (\textit{VN}), \textit{Chinese} (\textit{ZH}), and \textit{Thai} (\textit{TH}).}
  \label{fig:llama31_news2024}
\end{figure*}
\noindent\textbf{Setup.}
We fix the language set to
\begin{align}
    \mathcal{L} &= \{\text{EN},\, l_{\text{bridge}}\}
\end{align}
where
$l_{bridge} \in \{\text{FR},\, \text{DE},\, \text{NL},\, \text{VN},\, \text{ZH},\, \text{TH}\}$. We consider the case when \(|\mathcal{M}_1| = |\mathcal{M}_2| = 1\), with $\mathcal{M}_1$ and $\mathcal{M}_2$ containing
\textit{Llama} and \textit{Mistral} respectively. We also consider the three datasets: \textit{BookSum}, \textit{ScriptBase-alpha} , and \textit{News2024}. For each dataset \(\mathcal{D}\), every document \(d_i^{(0)} \in \mathcal{D}\) undergoes \(N = 100\) translation iterations with an iteration of the form:
\begin{equation}
\text{EN} \;\rightarrow\; l_{\text{bridge
}} \;\rightarrow\; \text{EN}.
\end{equation}
All translations within a single chain are performed by a single model.
Concretely, at iteration \(t\), the translation operator 
\begin{equation}
\mathcal{T}^{(t)} 
= \mathcal{T}_{\text{EN} \leftarrow l_{\text{bridge}}}^{m_1}
  \;\circ\;
  \mathcal{T}_{l_{\text{bridge}} \leftarrow \text{EN}}^{m_1}
\end{equation}
is applied to produce 
\begin{equation}
d_i^{(t)} = \mathcal{T}^{(t)}\bigl(d_i^{(t-1)}\bigr).
\label{eqn:d_i_t}
\end{equation}
This yields the sequence $(d_i^{(0)},\, d_i^{(1)},\, \ldots,\, d_i^{(100)})$ for each document \(d_i^{(0)} \in \mathcal{D}\).
\begin{hyp}[H\ref{hyp:first}]\label{hyp:first} We hypothesize that iterative translation chains better preserve relevance and factuality when the bridge language shares lexical overlap, script, and syntax with the source language. In contrast, languages markedly dissimilar from the source language are expected to introduce greater distortion over iterations. \end{hyp}
\noindent\textbf{Results.} 
Figure \ref{fig:llama31_news2024} presents \textit{Llama}'s iterative translation outcomes on the \textit{News2024} dataset. Across all language pairs, there is a gradual decline in both factuality and relevance. Notably, language pairs exclusively using Latin script—with bridge languages such as French, German, and Dutch—demonstrated superior preservation of these qualities compared to those employing non-Latin script bridge languages, which exhibited more pronounced distortions over successive iterations. A similar trend was observed for \textit{Llama} in the other datasets, while \textit{Mistral} showed an even more severe decline across all three datasets. Comprehensive results for the remaining datasets and models are provided in Appendix \ref{app:b1}.

The average gradient values of FActScore in Table~\ref{tab:avg_derivative_bsl} quantify the rate of factuality loss across translation iterations. For language pairs composed solely of Latin script languages, gradients remain close to zero across all datasets and LLMs, indicating minimal degradations. For instance, in the \textit{News2024} dataset, the average gradients for $\textbf{EN} \leftrightarrow \textbf{FR}$ are -0.004 (±0.003) with \textit{Llama} and -0.007 (±0.004) with \textit{Mistral}, while for $\textbf{EN} \leftrightarrow \textbf{DE}$ they are -0.005 (±0.003) and -0.011 (±0.006), respectively. In contrast, chains involving non-Latin scripts—particularly Thai—exhibit significantly faster factuality loss. In the \textit{BookSum} dataset, the $\textbf{EN} \leftrightarrow \textbf{TH}$ gradient is -0.026 (±0.014) with \textit{Llama} and -0.040 (±0.025) with \textit{Mistral}. This pattern is consistently observed across all evaluated language pairs, datasets, and models, with Thai demonstrating the highest rates of factual degradation.

\begin{table*}[ht]
    \centering
    {\fontsize{25}{35}\selectfont  
    \resizebox{\textwidth}{!}{%
    \begin{tabular}{ll cc cc cc cc cc cc}
    \toprule
     &  & \multicolumn{2}{c}{\textbf{EN $\leftrightarrow$ DE}} 
       & \multicolumn{2}{c}{\textbf{EN $\leftrightarrow$ FR}} 
       & \multicolumn{2}{c}{\textbf{EN $\leftrightarrow$ NL}} 
       & \multicolumn{2}{c}{\textbf{EN $\leftrightarrow$ TH}} 
       & \multicolumn{2}{c}{\textbf{EN $\leftrightarrow$ VN}} 
       & \multicolumn{2}{c}{\textbf{EN $\leftrightarrow$ ZH}} \\
    \cmidrule(lr){3-4} \cmidrule(lr){5-6} \cmidrule(lr){7-8} \cmidrule(lr){9-10} \cmidrule(lr){11-12} \cmidrule(lr){13-14}
    \textbf{Dataset} & \textbf{Model} 
       & \textbf{Avg Grad.} & \textbf{Std Err.} 
       & \textbf{Avg Grad.} & \textbf{Std Err.} 
       & \textbf{Avg Grad.} & \textbf{Std Err.} 
       & \textbf{Avg Grad.} & \textbf{Std Err.} 
       & \textbf{Avg Grad.} & \textbf{Std Err.} 
       & \textbf{Avg Grad.} & \textbf{Std Err.} \\
    \midrule
    \multirow{2}{*}{BookSum} 
       & \text{Llama} & -0.006 & 0.003 & \textbf{-0.005} & 0.003 & -0.006 & 0.003 & \underline{-0.026} & 0.014 & -0.009 & 0.005 & -0.021 & 0.011 \\
       & \text{Mistral} & -0.018 & 0.009 & \textbf{-0.014} & 0.007 & -0.016 & 0.008 & \underline{-0.040} & 0.025 & -0.031 & 0.018 & -0.028 & 0.015 \\
    \midrule
    \multirow{2}{*}{News2024} 
       &\text{Llama} & -0.005 & 0.003 & \textbf{-0.004} & 0.003 & -0.005 & 0.003 & \underline{-0.018} & 0.009 & -0.008 & 0.005 & -0.011 & 0.006 \\
       &\text{Mistral} & -0.011 & 0.006 & \textbf{-0.007} & 0.004 & -0.011 & 0.006 & \underline{-0.038} & 0.022 & -0.027 & 0.015 & -0.024 & 0.012 \\
    \midrule
    \multirow{2}{*}{ScriptBase-alpha} 
       &\text{Llama} & \textbf{-0.005} & 0.003 & -0.006 & 0.004 & -0.005 & 0.004 & \underline{-0.015} & 0.009 & -0.011 & 0.007 & -0.013 & 0.008 \\
       &\text{Mistral} & -0.010 & 0.006 & \textbf{-0.008} & 0.005 & -0.009 & 0.006 & \underline{-0.039} & 0.023 & -0.027 & 0.015 & -0.021 & 0.011 \\
    \bottomrule
    \end{tabular}%
    }
    }
    \caption{Comparison of average gradient and standard error values of FActScore for the different models and language pairs across datasets.}
    \label{tab:avg_derivative_bsl}
\end{table*}
\subsection{Experiment 2: Bilingual Two-player}
\label{sec:exp2} 

\noindent\textbf{Setup.} We fix the language set to
\begin{align}
    \mathcal{L} &= \{\text{EN},\, l_{\text{bridge}}\}
\end{align}
where $l_{bridge} \in \{\text{FR},\text{TH\}}$. Following the results presented in Section \ref{sec:exp1}, we selected \textbf{EN $\leftrightarrow$ FR} and \textbf{EN $\leftrightarrow$ TH} for Experiment 2, as they demonstrated the lowest and highest levels of information distortion, respectively. We consider a model set \(\mathcal{M}\) that includes both \textit{Llama} and \textit{Mistral}.

For this experiment, we used the \textit{News2024} dataset because, as shown in Section \ref{sec:exp1}, the choice of dataset did not significantly influence the observed trends, and to further mitigate data exposure \citep{luo-etal-2024-diverge, li2024latesteval}.

Unlike Experiment 1, where a single model was used for both translation directions, we allow each translation step to potentially use a different model. At iteration $t$, we define a two-component model sequence:\vspace{-8pt}
\begin{equation}
    \mathcal{M}^{(t)} = \bigl(m_1^{(t)}, m_2^{(t)}\bigr),
\end{equation}
where \(m_1^{(t)}\) is the model used for the translation from English to \(l_{\text{bridge}}\), and \(m_2^{(t)}\) is the model used for the translation from \(l_{\text{bridge}}\) to English. Each component is sampled uniformly from \(\mathcal{M}\).

The translation operator at iteration \(t\) is then defined as:
\begin{equation}
\mathcal{T}^{(t)} = \mathcal{T}_{\text{EN} \leftarrow l_{\text{bridge}}}^{m_{2}^{(t)}} \;\circ\; \mathcal{T}_{l_{\text{bridge}} \leftarrow \text{EN}}^{m_{1}^{(t)}}.
\end{equation}
The output document at iteration \(t\) is then determined as shown in Equation \ref{eqn:d_i_t}.
This yields the sequence 
$(d_i^{(0)},\, d_i^{(1)},\, \ldots,\, d_i^{(100)})$ for each document in the \textit{News2024} dataset.
\begin{hyp}[H\ref{hyp:second}]\label{hyp:second} We hypothesize that the coexistence of two different models in the same translation chain will add more distortions to the original information, thereby causing the original information to degrade over the successive iterations. \end{hyp}

\begin{figure*}[t]
    \centering
    \includegraphics[width=\textwidth]{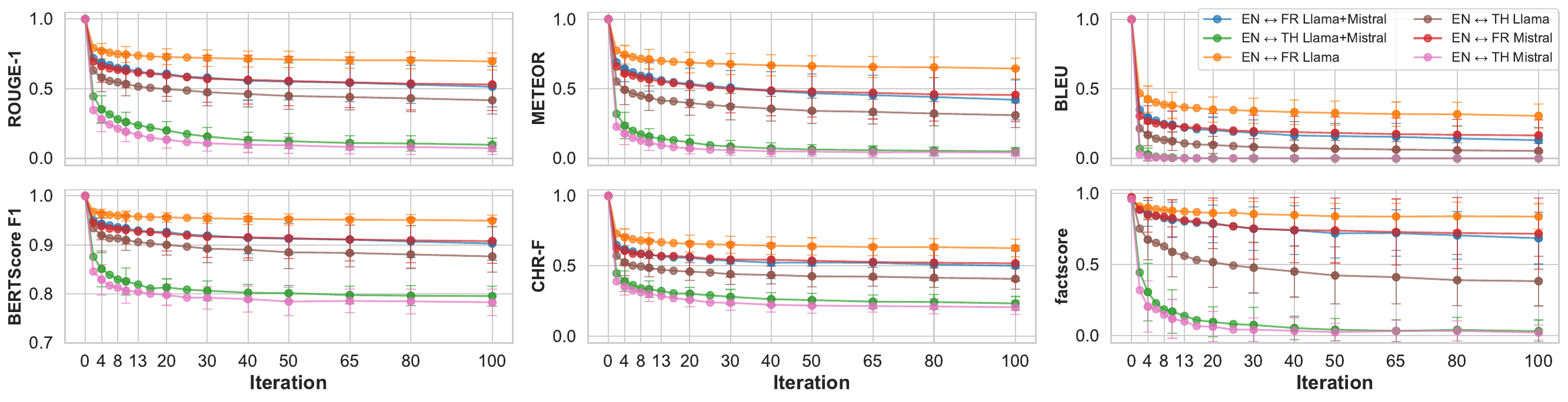}
    \caption{Comparison of metrics for the Bilingual Two-Player Experiment on the \textit{News2024} dataset, illustrating the interaction effects between \textit{Llama} and \textit{Mistral} on translation chains for 
    $\text{EN}\leftrightarrow \text{FR}$ and 
    $\text{EN}\leftrightarrow \text{TH}$, contrasted with their individual performances in the Bilingual Self-loop Experiment.}
    \label{fig:bmp_experiment}
\end{figure*}

\noindent\textbf{Results.} Figure~\ref{fig:bmp_experiment} shows distinct patterns in the collaborative performance of \textit{Llama} and \textit{Mistral} across different languages. In French, the joint translation chain did not enhance the preservation of factuality or textual relevance relative to the models operating independently; instead, the collaboration introduced additional distortions that further degraded all evaluation metrics. Conversely, in Thai, the collaboration of \textit{Llama} and \textit{Mistral} resulted in reduced distortion compared to \textit{Mistral} alone, though it still exhibited greater degradation than \textit{Llama} in isolation.

We further quantified the average gradient of FActScore. For French, the collaborative chain exhibited an average gradient of -0.007 (±0.004), confirming minimal factuality degradation, though slightly worse than the standalone performances of \textit{Llama} and \textit{Mistral}. In contrast, for Thai, the collaborative chain showed a lower average gradient of -0.035 (±0.019) when compared to the standalone chain of \textit{Mistral}. However, despite this lower decline, it was still outperformed by the standalone performance of \textit{Llama}, where factual degradation was less pronounced.
\subsection{Experiment 3: Multilingual Multiplayer}
\label{sec:exp3}

\noindent\textbf{Setup.} In this experiment, we design three settings of increasing complexity, each incorporating at least two bridge languages and at least two models within the same translation chain. The objective is to examine whether introducing a greater number of languages or models accelerates distortion.

\noindent\textbf{Setting 1.}  
We fix 
\begin{equation}
\mathcal{L} = \{\text{EN}, \text{FR}, \text{TH}\} \quad 
\label{eqn:languages_mm_setting1}
\end{equation}
and define $\mathcal{M}$ to contain both \textit{Llama} and \textit{Mistral}.

At each iteration \( t \), we sample a permutation \( \mathcal{L}^{(t)} = \pi^{(t)}(\mathcal{L}) \) that enforces a cyclic translation path:
\[
\text{EN} \rightarrow l_1^{(t)} \rightarrow l_2^{(t)} \rightarrow \text{EN},
\]
with \( l_1^{(t)} \) and \( l_2^{(t)} \) drawn from \(\{\text{FR}, \text{TH}\}\) and satisfying \( l_1^{(t)} \neq l_2^{(t)} \). The corresponding model sequence is
\begin{equation}
\mathcal{M}^{(t)} = \bigl(m_1^{(t)},\, m_2^{(t)},\, m_3^{(t)}\bigr),
\end{equation}
with each \( m_k^{(t)} \) sampled uniformly from \( \mathcal{M}\). The translation operator at iteration \(t\) is composed as:\vspace{-5pt}
\begin{equation}
\mathcal{T}^{(t)} = \mathcal{T}_{\text{EN} \leftarrow l_2^{(t)}}^{m_3^{(t)}} \circ \mathcal{T}_{l_2^{(t)} \leftarrow l_1^{(t)}}^{m_2^{(t)}} \circ \mathcal{T}_{l_1^{(t)} \leftarrow \text{EN}}^{m_1^{(t)}},
\label{eqn:mm_setting1_operator}
\end{equation}
which is applied iteratively to generate:  
\begin{equation}
d_i^{(t)} = \mathcal{T}^{(t)}\Bigl(d_i^{(t-1)}\Bigr).
\end{equation}
This produces \((d_i^{(0)},\, d_i^{(1)},\, \ldots,\, d_i^{(N)})\), where \(d_i^{(0)}\) is the original document and \(N = 100\).

\noindent\textbf{Setting 2.}  
We here retain $\mathcal{L}$ and the translation chain structure from Setting 1, utilizing the same translation operator as defined in Equation \ref{eqn:mm_setting1_operator}, while expanding $\mathcal{M}$ with an additional model, \textit{Gemma}, to assess the impact of adding more models of similar size into the chain.

\noindent\textbf{Setting 3.}  
We extend the language set to:
\begin{equation}
\mathcal{L} = \{\text{EN}, \text{FR}, \text{TH}, \text{ZH}, \text{DE}\},
\end{equation}
and hold $\mathcal{M}$ fixed from Setting 1.
The translation operator is then defined as: \vspace{-7pt} 
\begin{equation}
\mathcal{T}^{(t)} = \mathcal{T}_{\text{EN} \leftarrow l_4^{(t)}}^{m_5^{(t)}} \circ \cdots \circ \mathcal{T}_{l_1^{(t)} \leftarrow \text{EN}}^{m_1^{(t)}},
\end{equation}
applied to generate \(d_i^{(t)}\).
\begin{hyp}[H\ref{hyp:third}]\label{hyp:third} We hypothesize that higher complexity translation chains cause higher factual degradation of the source document. 
\end{hyp}

\noindent\textbf{Results.} 
As shown in Appendix \ref{app:b2},
all three experimental settings indicated a comparable degree of factual, lexical, and semantic degradation by the $100^{th}$ iteration across all evaluation metrics. However, differences emerged in the rate at which this degradation occurred. Specifically, Setting 3 exhibited the steepest decline in factual accuracy, with an average FActScore gradient of $-0.038 \pm 0.02$. By the $10^{th}$ generation, Setting 3's factuality had dropped to $0.054$, and by the $100^{th}$ generation, it further declined to $0.04$. Setting 1 followed closely, with an average gradient of $-0.036 \pm 0.02$, showing a factuality score of $0.063$ at the $10^{th}$ generation and $0.04$ at the $100^{th}$. Setting 2 exhibited the slowest rate of factual degradation, with an average gradient of $-0.034 \pm 0.02$, reaching $0.075$ at the $10^{th}$ generation and $0.04$ at the $100^{th}$.

\section{Ablation Studies}
\begin{figure*}[t]
    \centering
    \includegraphics[width=\textwidth]{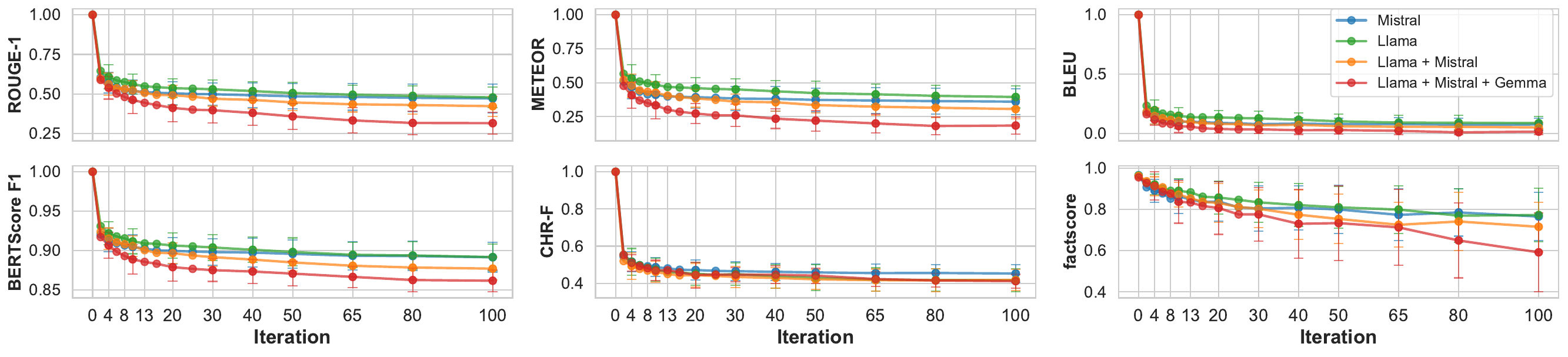}
    \caption{Results of rephrasing experiments on 30 randomly sampled documents from the \textit{News2024} dataset. The figure compares the performance of individual models (\textit{Llama}, \textit{Mistral}, and \textit{Gemma}) and their collaborative combinations over 100 rephrasing iterations.}
    \label{fig:rephrase_ablation}
\end{figure*}
\begin{figure*}[t]
    \centering
    \includegraphics[width=\textwidth]{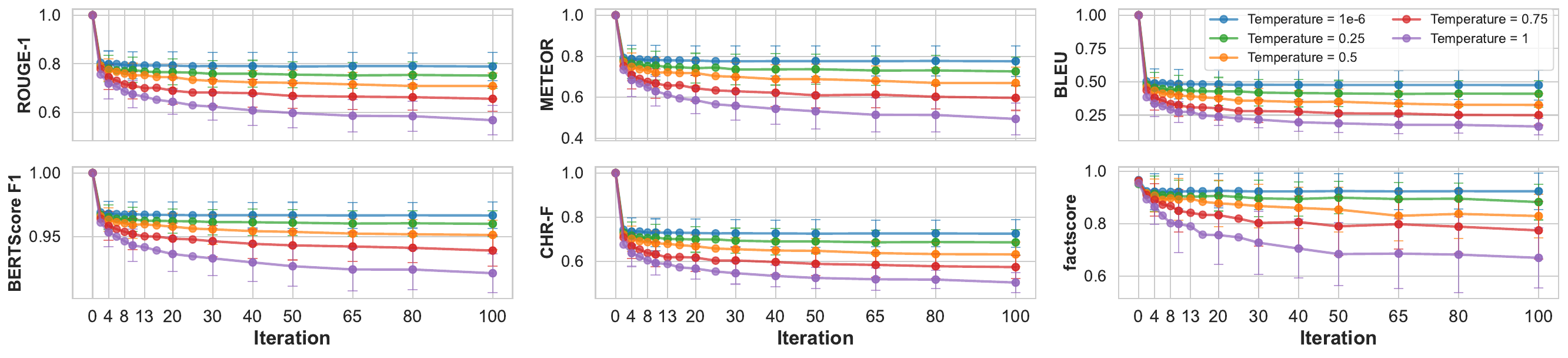}
    \caption{Impact of temperature variation on \textit{Llama} outputs for 30 randomly sampled documents from the \textit{News2024} dataset, evaluated on the $\text{EN} \leftrightarrow \text{FR}$ translation chain.}
    \label{fig:temperature_variation}
\end{figure*}
\begin{figure*}[t]
    \centering
    \includegraphics[width=\textwidth]{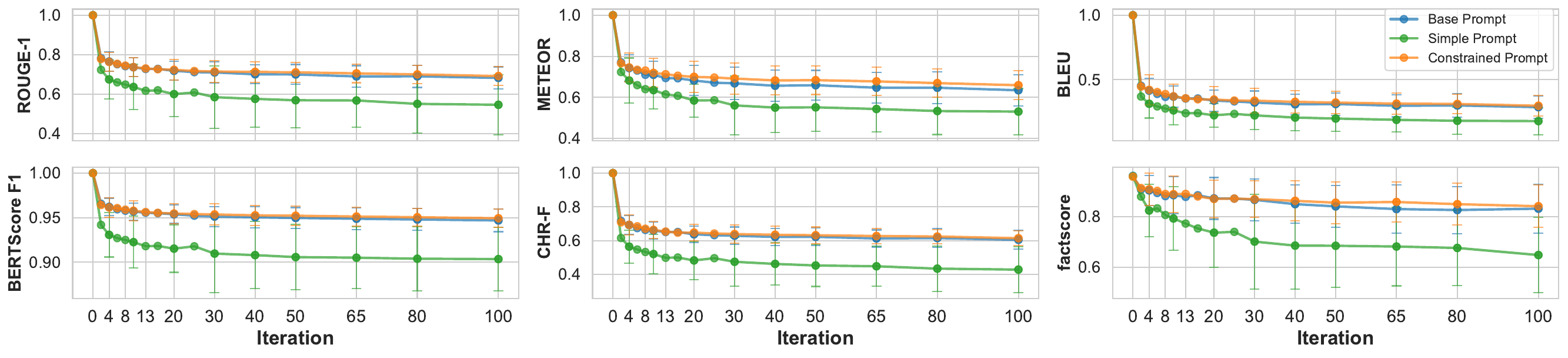}
    \caption{Impact of the prompt's level of constraint on noise accumulation measure on each of the metrics for \textit{Llama} on 30 randomly sampled documents from the \textit{News2024} dataset in iterative translation}
    \label{fig:prompt_ablation}
\end{figure*}
\subsection{Other Tasks: Rephrasing}
\label{sec:rephrasing_sec}
Building on our findings in section \ref{sec:experiments}, we extend our experiments to explore whether information distortion manifests in other types of iterative generation chains. Inspired by the work of \citet{perez2024llms}, who examined the evolution of toxicity, positivity, difficulty, and length in rephrasing as well as in continuation and inspiration-taking chains, we further probe the effects of information distortion in more complex rephrasing chains.

In this task, the model is instructed to rephrase a given document while preserving its full meaning (the full rephrasing prompt can be found in Appendix \ref{sec:prompts}). We randomly sampled 30 documents from \textit{News2024} and conducted four experiments based on the setups from Sections \ref{sec:exp1}, \ref{sec:exp2}, and \ref{sec:exp3} (Setting 2). These experiments tested standalone rephrasing chains, the collaborative effects of \textit{Llama} and \textit{Mistral}, and an extended setup incorporating \textit{Gemma} into the chain.

Rephrasing results are presented in Figure~\ref{fig:rephrase_ablation}. Textual relevance metrics reveal rapid degradation of lexical and semantic properties over iterations. Among individual models, \textit{Llama} shows the slowest divergence in textual relevance, with \textit{Mistral} following. When these models collaborate, the degradation in textual relevance increases, and combining \textit{Llama}, \textit{Mistral}, and \textit{Gemma} leads to the steepest decline, particularly after 100 iterations. The same order was observed when evaluating factuality, although the loss was steadier, without clear convergence at the $100^{th}$ iteration.

\subsection{Temperature Variation Affects Outputs}
To further investigate the impact of decoding parameters on the models' outputs, we conducted several experiments using \textit{Llama} across a spectrum of temperature parameter values, including \(1 \times 10^{-6}\), 0.25, 0.5, 0.75, and 1.0 on 30 randomly sampled documents from \textit{News2024}.

From Figure \ref{fig:temperature_variation}, higher temperature settings lead to greater factual and semantic degradation. At extremely low temperatures (\(1 \times 10^{-6}\)), factuality drops slightly in the first two iterations but stabilizes thereafter. As temperature increases, stability diminishes, and factuality gradually diverges. Higher temperatures exacerbate this trend, with maximum temperature (1.0) causing the steepest decline, showing continuous divergence. Additional examples can be found in Appendix \ref{sec:examples analysis}. \vspace{-5pt}
\subsection{Sensitivity of Iterative Translation Outputs to the Chosen Prompt}\vspace{-2pt}
We subsequently investigated the influence of the translation prompt on the outputs produced by the iterative process. To this end, 30 documents were randomly sampled from the \textit{News2024} dataset, and \textit{Llama} was tasked with translating them using three distinct prompts characterized by varying levels of constraint: simple, base (used in all our experiments), and constrained. The complete prompts are provided in Appendix \ref{sec:prompts}.

Figure \ref{fig:prompt_ablation} illustrates that the level of constraint imposed by the prompt markedly affects the model's generation. Specifically, more constrained prompts were found to result in higher levels of relevance and factuality preservation.


\section{Discussion and Conclusion}
As LLMs increasingly shape online content, the likelihood that they re-process their own outputs continues to rise. This study confirms that such iterative generation leads to progressive information distortion, akin to the ``broken telephone'' effect in human communication. Our findings from translation-based experiments are multifaceted.

\medskip

\noindent\textbf{Effect of intermediate language(s) on information distortion.} As found in Experiment 1, different language chains have varying levels of sensitivity to information distortion. As presented in Figure \ref{fig:llama31_news2024}, we found that transmitting information between English and a highly similar language significantly reduces the distortion effect, while transmitting through a dissimilar language results in a more pronounced distortion. We suggest that this variation in information retention and distortion stems from the proportion of each language encountered during the models' training, with underrepresented languages experiencing greater distortion.

\medskip

\noindent\textbf{Chains of higher complexity may result in higher levels of distortion.} Experiments 2 and 3 showed that increasing the levels of complexity of chains can result in higher levels of distortion. Figure \ref{fig:bmp_experiment} illustrates how the combination of \textit{Llama} and \textit{Mistral} amplified the distortion in the chain when French served as the bridge language. However, when Thai was used as the bridge language, their collaboration helped reduce distortion—likely due to the stronger model (\textit{Llama}) and the weaker model (\textit{Mistral}) interacting with an intermediate language that may have been underrepresented in \textit{Mistral}'s training compared to \textit{Llama}. Moreover, we observed that increasing the number of languages in the translation chain amplifies information distortion, likely due to the cumulative effects of longer generation sequences. In contrast, incorporating \textit{Gemma} into the chain improved information retention, which we hypothesize stems from its larger parameter count—one to two billion more than \textit{Llama} and \textit{Mistral}. We leave the broader impact of model scaling for future work. 

\medskip

\noindent\textbf{Information distortion can be reduced through temperature control and constrained prompting.} Our findings suggest that while information distortion is unavoidable, it can be significantly mitigated through careful control of the model's generation temperature. Figure \ref{fig:temperature_variation} shows that higher temperature values lead to greater distortion in the outputs, which we attribute to increased model creativity. A higher temperature encourages the generation of atypical tokens that may not fully preserve the meaning of the source document. Additionally, our analysis of prompt effects revealed that less constrained prompts contribute to greater noise accumulation over multiple iterations, resulting in higher divergence from the original meaning.

\medskip

These findings underscore the need for strategies to mitigate such degradation and ensure the reliability of AI-generated content.

\section*{Limitations}
While our study utilizes datasets from three distinct domains—book summaries, movie scripts, and news articles—these sources share similar characteristics and may not reflect the rare or long-tailed information found in specialized domains. Moreover, due to computational resource limitations, our experiments are restricted to models with 7–9 billion parameters, using their default generation settings. Future work should investigate whether incorporating datasets from specialized domains, employing larger models, or varying generation strategies (e.g., greedy decoding) impacts the degree of information distortion in iterative generation chains.

\bibliography{googlescholar,aclanthology}

\onecolumn
\appendix
\section{Additional Details}
All experiments were conducted using NVIDIA A100 (40GB VRAM) and A10 (24GB VRAM) GPU clusters. The compute allocation totaled 54 GPU-days, comprising 36 GPU-days on 8×A100 nodes and 18 GPU-days on 4×A10 nodes.

\section{Experiments Results Visualizations}

We hereby present the complement of the visualizations of results from sections \ref{sec:exp1} and \ref{sec:exp3}

\subsection{Experiment 1: Bilingual Self-loop} \label{app:b1}

\subsubsection{Llama}
\begin{figure*}[ht]
    \centering
    \includegraphics[width=\textwidth]{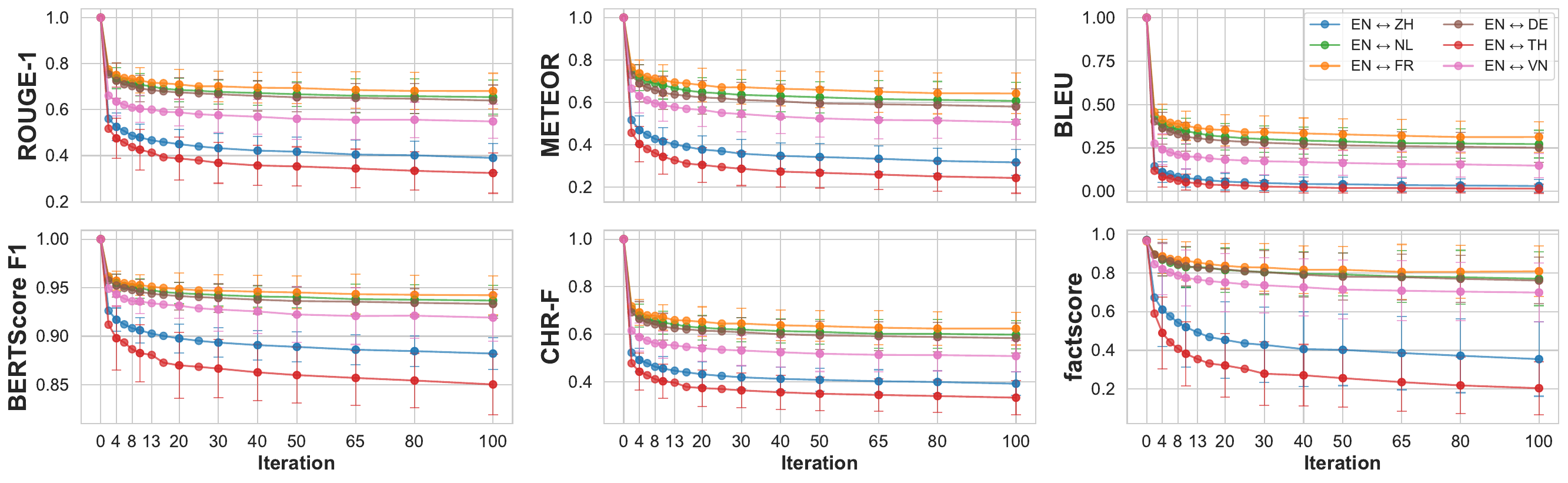}
    \caption{Results of \textit{Llama} in the Bilingual Self-loop Experiment showing metrics evolution across translation iterations over the \textit{BookSum} dataset for \textit{French} (\textit{FR}), \textit{German} (\textit{DE}), \textit{Dutch} (\textit{NL}), \textit{Vietnamese} (\textit{VN}), \textit{Chinese} (\textit{ZH}), and \textit{Thai} (\textit{TH})}
    \label{fig:llama31_booksum}
\end{figure*}

\begin{figure*}[ht]
    \centering
    \includegraphics[width=\textwidth]{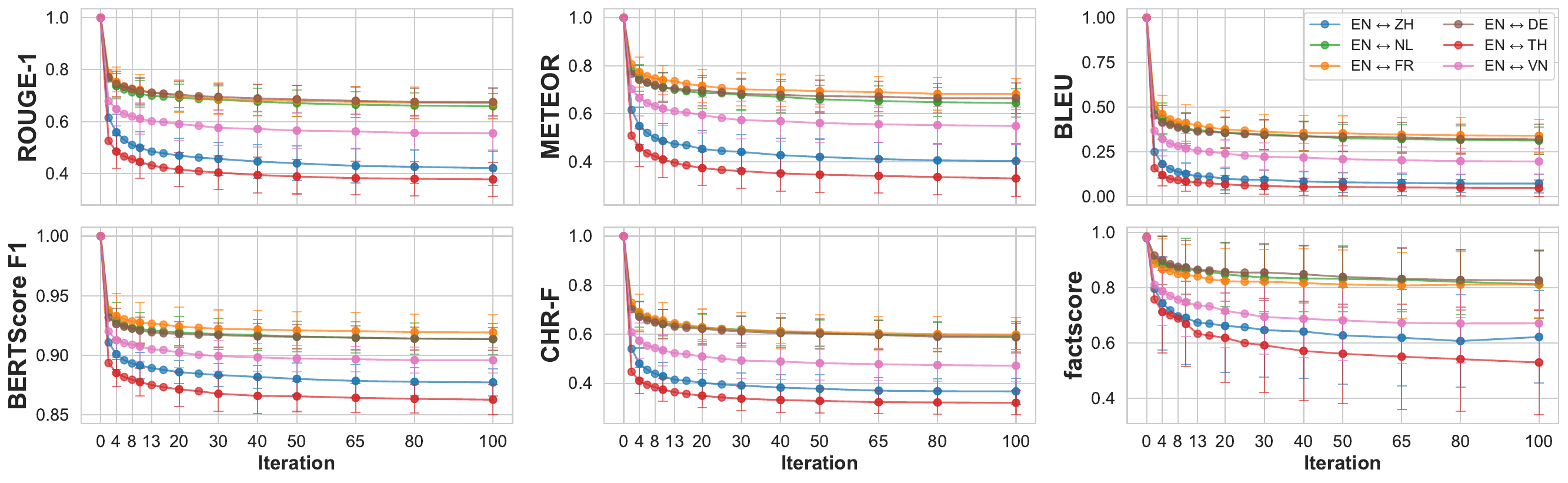}
    \caption{Results of \textit{Llama} in the Bilingual Self-loop Experiment showing metrics evolution across translation iterations over the \textit{ScriptBase-alpha} dataset for \textit{French} (\textit{FR}), \textit{German} (\textit{DE}), \textit{Dutch} (\textit{NL}), \textit{Vietnamese} (\textit{VN}), \textit{Chinese} (\textit{ZH}), and \textit{Thai} (\textit{TH})}
    \label{fig:llama31_scriptbase}
\end{figure*}
\newpage
\subsubsection{Mistral}
\begin{figure*}[ht]
    \centering
    \includegraphics[width=\textwidth]{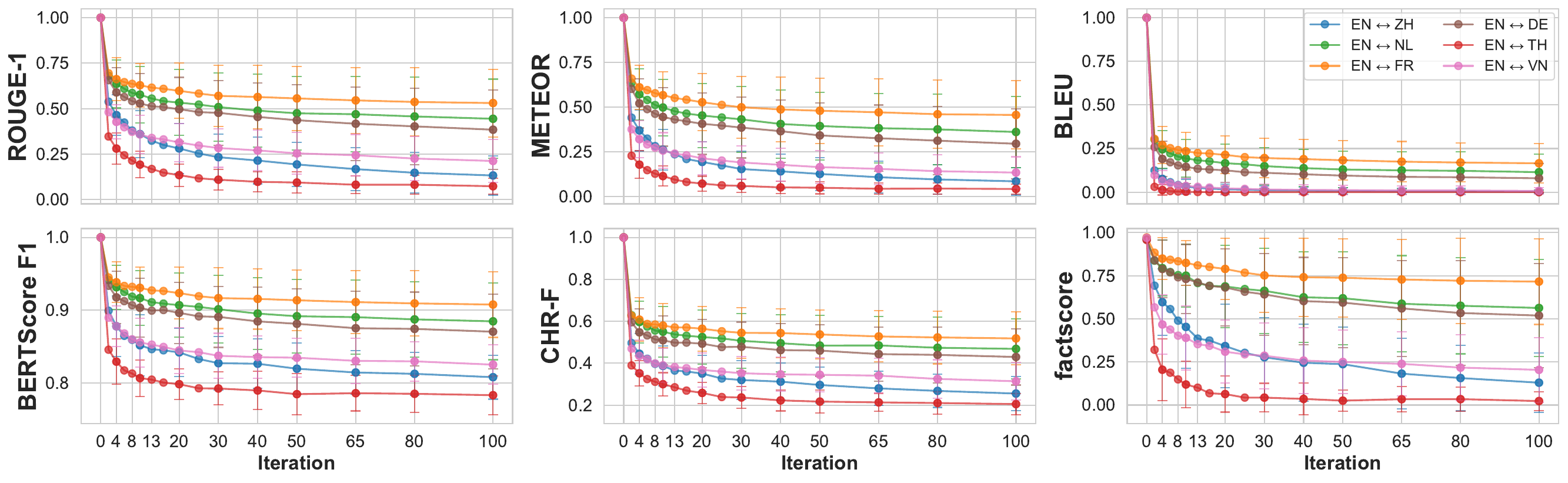}
    \caption{Results of \textit{Mistral} in the Bilingual Self-loop Experiment showing metrics evolution across translation iterations over the \textit{News2024} dataset for \textit{French} (\textit{FR}), \textit{German} (\textit{DE}), \textit{Dutch} (\textit{NL}), \textit{Vietnamese} (\textit{VN}), \textit{Chinese} (\textit{ZH}), and \textit{Thai} (\textit{TH})}
    \label{fig:mistral7bv2_news2024}
\end{figure*}
\begin{figure*}[ht]
    \centering
    \includegraphics[width=\textwidth]{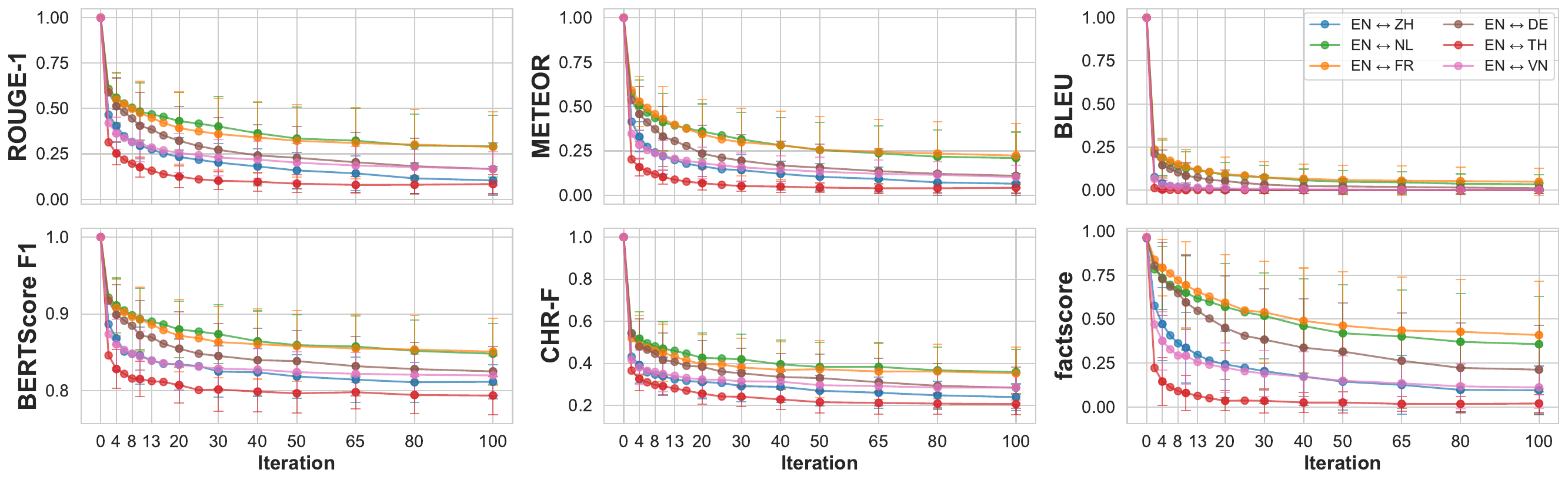}
    \caption{Results of \textit{Mistral} in the Bilingual Self-loop Experiment showing metrics evolution across translation iterations over the \textit{BookSum} dataset for \textit{French} (\textit{FR}), \textit{German} (\textit{DE}), \textit{Dutch} (\textit{NL}), \textit{Vietnamese} (\textit{VN}), \textit{Chinese} (\textit{ZH}), and \textit{Thai} (\textit{TH})}
    \label{fig:mistral7bv2_booksum}
\end{figure*}
\begin{figure*}[ht]
    \centering
    \includegraphics[width=\textwidth]{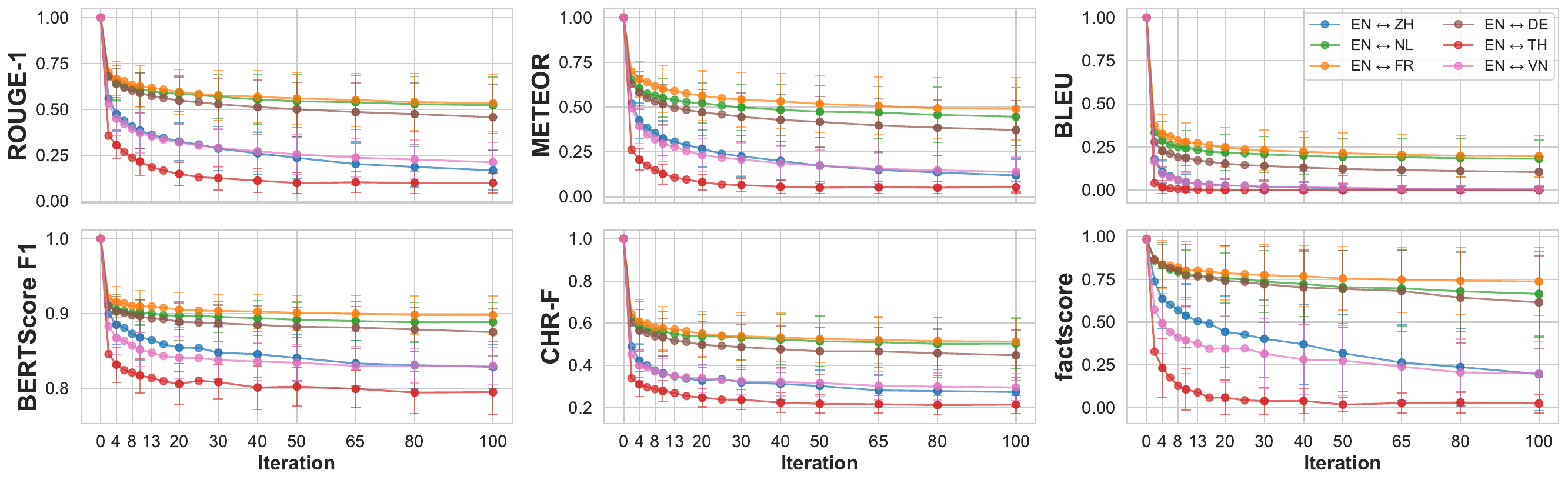}
    \caption{Results of \textit{Mistral} in the Bilingual Self-loop Experiment showing metrics evolution across translation iterations over the \textit{ScriptBase-alpha} dataset for \textit{French} (\textit{FR}), \textit{German} (\textit{DE}), \textit{Dutch} (\textit{NL}), \textit{Vietnamese} (\textit{VN}), \textit{Chinese} (\textit{ZH}), and \textit{Thai} (\textit{TH})}
    \label{fig:mistral7bv2_scriptbase}
\end{figure*}

\newpage

\subsection{Experiment 3: Multilingual Multiplayer} \label{app:b2}

\begin{figure*}[ht]
    \centering
    \includegraphics[width=\textwidth]{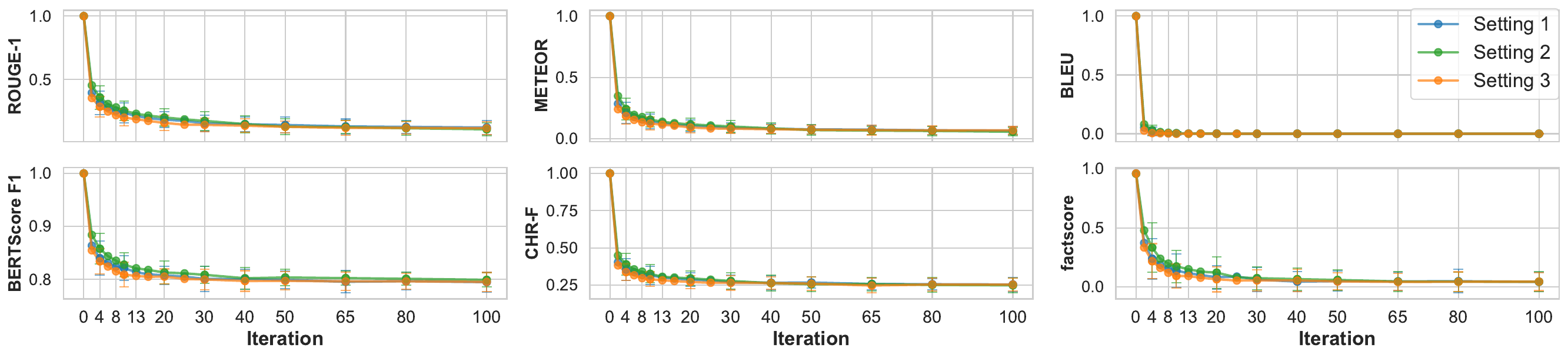}
    \caption{Results of Experiment 3 showing metrics evolution across translation iterations over the \textit{News2024} dataset. \textbf{\textit{Setting 1}} presents the translation chain of English with the bridge languages French and Thai. \textbf{\textit{Setting 2}} maintains the same chain as \textit{Setting 1}, while adding \textit{Gemma} to the set of models. \textbf{\textit{Setting 3} }further add the bridge languages Chinese and German to the chain from Setting 1.}
    \label{fig:mm_experiment}
\end{figure*}
\newpage

\newpage
\section{Generation Settings and Decoding Parameters}
\label{app:generation_settings}

Each model was used for inference with its default decoding parameters as specified in their respective official documentation. We capped the maximum number of newly generated tokens at 8000 to encourage open-ended generation and ensure translations could conclude naturally. The choice of default parameters was made to reflect common practical usage scenarios where models are often deployed with these settings. The default sampling-based decoding parameters used for each model in our experiments are detailed in Table \ref{tab:decoding_params}.

\begin{table}[H] 
    \centering
    \small
    \begin{tabular}{@{}lcc@{}}
        \toprule
        Model                     & Temperature & Top-p \\
        \midrule
        Llama-3.1-8B-Instruct     & 0.6         & 0.9   \\
        Mistral-7B-Instruct-v0.2  & 0.0  & N/A   \\ 
        Gemma-2-9B-it             & 1.0         & 0.95  \\
        \bottomrule
    \end{tabular}
    \caption{Default decoding parameters used for \textit{LLama}, \textit{Mistral}, and \textit{Gemma}.}
    \label{tab:decoding_params}
\end{table}
Further exploration of diverse decoding strategies, such as greedy decoding or beam search, remains an avenue for future work.

\section{Additional BLEURT Score Evaluations}
\label{app:bleurt_scores}

To further reinforce our findings, we computed BLEURT scores \citep{sellam-etal-2020-bleurt}—known for their high correlation with human judgments—across the full outputs of our primary experiments and rephrasing tasks. While main evaluations assessed FActScore for factual preservation alongside traditional textual relevance metrics (BLEU, ROUGE, METEOR, BERTScore) to capture changes in linguistic form over successive iterations, the BLEURT scores presented below (Tables \ref{tab:bleurt_bsl} to \ref{tab:bleurt_rephrase}) further substantiate our conclusions regarding information distortion. Higher BLEURT scores indicate better quality and closer semantic similarity to the original text.

\subsection{Bilingual Self-loop BLEURT Scores}
Table \ref{tab:bleurt_bsl} presents the BLEURT scores for the Bilingual Self-loop experiment, corresponding to the \textit{Llama} model on the \textit{News2024} dataset (as detailed in Section \ref{sec:exp1} and visualized for other metrics in Figure \ref{fig:llama31_news2024}).

\begin{table*}[ht]
    \centering
    \scriptsize

    \begin{tabular}{@{}lrrrrrrrrrrrr@{}}
        \toprule
        Language Pair/ & \multicolumn{12}{c}{Iteration} \\
        \cmidrule(lr){2-13}
        Iteration & 0 & 3 & 6 & 10 & 15 & 20 & 30 & 40 & 50 & 65 & 80 & 100 \\
        \midrule
        EN $\leftrightarrow$ FR & 0.949 & 0.727 & 0.714 & 0.704 & 0.696 & 0.693 & 0.688 & 0.683 & 0.678 & 0.676 & 0.675 & 0.670 \\
        EN $\leftrightarrow$ NL & 0.949 & 0.721 & 0.704 & 0.689 & 0.679 & 0.670 & 0.664 & 0.657 & 0.654 & 0.649 & 0.645 & 0.642 \\
        EN $\leftrightarrow$ DE & 0.949 & 0.711 & 0.688 & 0.677 & 0.669 & 0.660 & 0.654 & 0.645 & 0.641 & 0.635 & 0.630 & 0.625 \\
        EN $\leftrightarrow$ VN & 0.949 & 0.681 & 0.659 & 0.643 & 0.632 & 0.625 & 0.614 & 0.606 & 0.603 & 0.596 & 0.589 & 0.586 \\
        EN $\leftrightarrow$ ZH & 0.949 & 0.618 & 0.587 & 0.570 & 0.551 & 0.537 & 0.522 & 0.514 & 0.508 & 0.498 & 0.494 & 0.489 \\
        EN $\leftrightarrow$ TH & 0.950 & 0.584 & 0.537 & 0.514 & 0.496 & 0.482 & 0.462 & 0.454 & 0.447 & 0.440 & 0.434 & 0.426 \\
        \bottomrule
    \end{tabular}
    \caption{BLEURT scores for the Bilingual Self-loop experiment using \textit{Llama} on the \textit{News2024} dataset. Scores show the evolution of text quality over 100 iterations for different language pairs.}
\label{tab:bleurt_bsl}
\end{table*}

Similar to our previous results for textual relevance and factuality in this experimental setup, the BLEURT scores exhibit the same trends. Specifically, there is a consistent decline in scores across all iterations, indicating progressive degradation of text quality. This degradation is less severe for language pairs where the bridge language uses a Latin script and shares more similarities with English (e.g., EN $\leftrightarrow$ FR, EN $\leftrightarrow$ NL, EN $\leftrightarrow$ DE), which show higher BLEURT scores throughout the iterations compared to pairs involving non-Latin scripts or more distant languages (e.g., EN $\leftrightarrow$ VN, EN $\leftrightarrow$ ZH, and particularly EN $\leftrightarrow$ TH). This observation aligns with Hypothesis \ref{hyp:first}, which posited that lexical and script similarity would influence the degree of information distortion.
\subsection{Bilingual Two-Player BLEURT Scores}
Table \ref{tab:bleurt_bt} shows BLEURT scores for the Bilingual Two-Player experiment on the \textit{News2024} dataset (detailed in Section \ref{sec:exp2} and Figure \ref{fig:bmp_experiment}), involving \textit{Llama} and \textit{Mistral}.

\begin{table*}[htbp]
    \centering
    \scriptsize
    \begin{tabular}{@{}lrrrrrrrrrrrr@{}}
        \toprule
        Language Pair/ & \multicolumn{12}{c}{Iteration} \\
        \cmidrule(lr){2-13}
        Iteration & 0 & 3 & 6 & 10 & 15 & 20 & 30 & 40 & 50 & 65 & 80 & 100 \\
        \midrule
        EN $\leftrightarrow$ FR & 0.949 & 0.684 & 0.661 & 0.640 & 0.622 & 0.614 & 0.597 & 0.592 & 0.575 & 0.570 & 0.564 & 0.547 \\
        EN $\leftrightarrow$ TH & 0.949 & 0.392 & 0.351 & 0.330 & 0.327 & 0.318 & 0.288 & 0.284 & 0.270 & 0.271 & 0.262 & 0.264 \\
        \bottomrule
    \end{tabular}
        \caption{BLEURT scores for the Bilingual Two-Player experiment on the \textit{News2024} dataset.}
    \label{tab:bleurt_bt}
\end{table*}

\subsection{Multilingual Multiplayer BLEURT Scores}
Table \ref{tab:bleurt_mm} shows BLEURT scores for the Multilingual Multiplayer experiment on the \textit{News2024} dataset (detailed in Section \ref{sec:exp3} and Appendix \ref{app:b2}, Figure \ref{fig:mm_experiment}).

\begin{table*}[htbp]
    \centering
    \scriptsize
    \begin{tabular}{@{}lrrrrrrrrrrrr@{}}
        \toprule
        Setting / & \multicolumn{12}{c}{Iteration} \\
        \cmidrule(lr){2-13}
        Iteration & 0 & 3 & 6 & 10 & 15 & 20 & 30 & 40 & 50 & 65 & 80 & 100 \\
        \midrule
        Setting 1 & 0.949 & 0.385 & 0.348 & 0.334 & 0.333 & 0.329 & 0.311 & 0.295 & 0.305 & 0.304 & 0.302 & 0.295 \\
        Setting 2 & 0.949 & 0.414 & 0.360 & 0.330 & 0.328 & 0.312 & 0.312 & 0.292 & 0.284 & 0.275 & 0.274 & 0.283 \\
        Setting 3 & 0.949 & 0.397 & 0.350 & 0.326 & 0.312 & 0.307 & 0.309 & 0.303 & 0.298 & 0.296 & 0.287 & 0.300 \\
        \bottomrule
    \end{tabular}
        \caption{BLEURT scores for the Multilingual Multiplayer experiment on the \textit{News2024} dataset.}
    \label{tab:bleurt_mm}
\end{table*}

\subsection{Rephrasing Task BLEURT Scores}
Table \ref{tab:bleurt_rephrase} shows BLEURT scores for the rephrasing task on the \textit{News2024} dataset (detailed in Section \ref{sec:rephrasing_sec} and Figure \ref{fig:rephrase_ablation}).

\begin{table*}[htbp]
    \centering
    \scriptsize
    \begin{tabular}{@{}lrrrrrrrrrrrr@{}}
        \toprule
        Model Combination / & \multicolumn{12}{c}{Iteration} \\
        \cmidrule(lr){2-13}
        Iteration & 0 & 3 & 6 & 10 & 15 & 20 & 30 & 40 & 50 & 65 & 80 & 100 \\
        \midrule
        Llama                 & 0.945 & 0.645 & 0.619 & 0.602 & 0.591 & 0.587 & 0.580 & 0.567 & 0.557 & 0.551 & 0.542 & 0.537 \\
        Mistral               & 0.945 & 0.593 & 0.577 & 0.562 & 0.552 & 0.545 & 0.538 & 0.535 & 0.533 & 0.533 & 0.528 & 0.521 \\
        Llama + Mistral       & 0.945 & 0.614 & 0.585 & 0.572 & 0.558 & 0.548 & 0.538 & 0.518 & 0.512 & 0.502 & 0.492 & 0.490 \\
        Llama + Mistral + Gemma & 0.945 & 0.587 & 0.552 & 0.531 & 0.507 & 0.496 & 0.480 & 0.465 & 0.457 & 0.443 & 0.438 & 0.427 \\
        \bottomrule
    \end{tabular}
    \caption{BLEURT scores for the rephrasing task on the \textit{News2024} dataset.}
\label{tab:bleurt_rephrase}
\end{table*}

These BLEURT evaluations, due to their high alignment with human judgment, reinforce our original analysis and confirm that the pattern of iterative degradation observed in our experiments is robust. The decline in BLEURT scores over iterations across different tasks and settings provides further evidence for the "broken telephone" effect in LLM-based iterative generation.

\section{Generation Prompts}
\label{sec:prompts}
\begin{figure}[H]
  \centering
  \hrule\vspace{1em}
  \small
  \begin{verbatim}
You are a translation expert. Given a passage, a source language, and a target language, 
translate the passage from the source language to the target language while preserving
all the original meaning and without losing any context. 
Do not write an introduction or a summary. Return only the translated passage.
Translate the following text from {source_language} to {target_language}: {document}
  \end{verbatim}
  \hrule\vspace{10pt}
  \caption{Base (main) translation Prompt for the translation chains experiments}
\label{fig:translation_prompt}
\end{figure}\vspace{-30pt}
\begin{figure}[H]
  \centering
  \hrule\vspace{1em}
  \small
  \begin{verbatim}
Translate the following text from {source_language} to {target_language}: {document}
  \end{verbatim}
  \hrule\vspace{10pt}
  \caption{Simple translation Prompt for the prompt ablation}
\label{fig:simple_prompt}
\end{figure}\vspace{-30pt}
\begin{figure}[H]
  \centering
  \hrule\vspace{1em}
  \small
  \begin{verbatim}
You are a translation expert. Please follow these instructions carefully:
- Task: You will receive a paragraph in {source_language}.
- Objective: Translate the paragraph into {target_language}.
- Guidelines:
  - Do not write an introduction or a summary. 
  - Preserve the original meaning entirely; ensure no information is lost or altered.
  - Do not add, omit, or modify any details from the original paragraph.
  - Maintain the tone and style as closely as possible.
- Paragraph: {document}
  \end{verbatim}
  \hrule\vspace{10pt}
  \caption{Constrained translation Prompt for the prompt ablation}
  \label{fig:constrained_prompt}
\end{figure}\vspace{-30pt}
\begin{figure}[H]
  \centering
  \hrule\vspace{1em}
  \small
  \begin{verbatim}
Given a passage, rephrase it while preserving all the original meaning and
without losing any context. 
Do not write an introduction or a summary. Return only the rephrased passage.

Rephrase the following text: {document}
  \end{verbatim}
  \hrule\vspace{10pt}
  \caption{Rephrasing prompt used for the rephrasing task}
  \label{fig:rephrase_prompt}
\end{figure}
\section{Examples Analysis}
\label{sec:examples analysis}
\begin{table*}[ht]
    \centering
    \begin{tabular}{|p{0.9\textwidth}|}
        \hline
        \small \textbf{Temperature 1e-16:} \namecolor{UEFA} has imposed \finecolor{fines} on the \namecolor{English Football Association} and the \namecolor{Football Association of Ireland} after their \damagecolor{national anthems} were \outcomecolor{booed} before \locationcolor{Ireland played England} in the \locationcolor{Nations League} in September. \\
        \hline
        \small \textbf{Temperature 0.25:} The \namecolor{Union of European Football Associations (UEFA)} has imposed \finecolor{fines} on the \namecolor{England Football Association} and the \namecolor{Football Association of Ireland} after their \damagecolor{national anthems} were \outcomecolor{booed} before \locationcolor{Ireland played England} in the \locationcolor{Nations League} in September. \\
        \hline
        \small \textbf{Temperature 0.5:} The \namecolor{UEFA Football Federation} has \finecolor{sanctioned} the \namecolor{England Football Association} and the \namecolor{Football Association of Ireland} after their \damagecolor{national anthems} were \outcomecolor{insulted} before \locationcolor{Ireland played England} in the \locationcolor{Nations League} in September. \\
        \hline
        \small \textbf{Temperature 0.75:} The \namecolor{UEFA Football Federation} has \finecolor{sanctioned} the \namecolor{Football Association (FA) of England} and the \namecolor{Football Association of Ireland (FAI)} after their \damagecolor{national anthems} were \outcomecolor{deemed offensive} before the \locationcolor{UEFA Nations League match} between \locationcolor{England and Ireland} in September. \\
        \hline
        \small \textbf{Temperature 1:} The \namecolor{Football Association (FA) of England} and the \namecolor{Football Association of Ireland (FAI)} have been \finecolor{sanctioned} by the \namecolor{Union of European Football Associations (UEFA)} due to \newdetailcolor{incidents} that occurred prior to the \locationcolor{UEFA Nations League match} between \locationcolor{England and Ireland} in September. \\
        \hline
    \end{tabular}
    \caption{An example of a news article highlighting the effect of temperature variation on the iterative translation process with English as the source language and French as the bridge language using \textit{Llama} after 100 iterations. Color key: \namecolor{Entities}, \finecolor{Financial Actions}, \damagecolor{Cultural Elements}, \outcomecolor{Controversial Outcomes}, \locationcolor{Events/Locations}, \newdetailcolor{Emergent Details}.}
    \label{table_temperature}
\end{table*}
\normalsize\vspace{-10pt}
\end{document}